# Automatically extracting, ranking and visually summarizing the treatments for a disease

**Prakash Reddy Putta, B.Tech[1,2], John J. Dzak III, BS[1], Siddhartha R. Jonnalagadda, PhD[1]**
[1]Division of Health and Biomedical Informatics, Northwestern University Feinberg School of Medicine, Chicago, IL
[2]Department of Management and Information Systems, University of Illinois, Chicago, IL

**Abstract**

*Clinicians are expected to have up-to-date and broad knowledge of disease treatment options for a patient. Online health knowledge resources contain a wealth of information. However, because of the time investment needed to disseminate and rank pertinent information, there is a need to summarize the information in a more concise format. Our aim of the study is to provide clinicians with a concise overview of popular treatments for a given disease using information automatically computed from Medline abstracts. We analyzed the treatments of two disorders – Atrial Fibrillation and Congestive Heart Failure. We calculated the precision, recall, and f-scores of our two ranking methods to measure the accuracy of the results. For Atrial Fibrillation disorder, maximum f-score for the New Treatments weighing method is 0.611, which occurs at 60 treatments. For Congestive Heart Failure disorder, maximum f-score for the New Treatments weighing method is 0.503, which occurs at 80 treatments.*

**Introduction**

Clinicians are expected to have up-to-date and broad knowledge of disease treatment options for a patient. Online health knowledge resources contain a wealth of information. However, because of the time investment needed to disseminate and rank pertinent information, there is a need to summarize the information in a more concise format. In a 1985 seminal study, Covell et al.[1] observed that physicians raised two questions for every three patients seen in an outpatient setting. In 70% of the cases, these questions were not answered. More recent research[2] has produced similar results, with little improvement compared to Covell et al.'s findings.

Based on Ely et al.'s taxonomy and their studies with clinicians at the point of care[2-4], the major information types needed at the point of care are treatment and diagnosis. Treatment questions (drugs and other therapy) account for about 44% of the questions asked and diagnosis-related questions constitute about 38% of the questions asked. The goal of this research is to provide clinicians with a concise overview of different treatments for a given disease. Given this information, a clinician can efficiently research common or emerging treatments while skipping over the unimportant treatments. In this study, we explored the feasibility of extracting treatments from Medline abstracts for 'Congestive Heart Failure' and 'Atrial Fibrillation' and providing an interactive display of the treatments through a web-interface.

Medline abstracts have been used recently to answer clinical questions by extracting individual sentences[5]. They have also been used to graphically display treatments and disorders as nodes in a graph, where the salient treatments are found using network centrality measures[6]. In this paper, we are studying an alternative format of summarizing therapeutic options where the clinician users can enter their disorder of interest in natural language or code, see the treatments ranked by frequency across different year ranges (although they might even be ranked using network centrality measures across different year ranges), and also compare different treatments or combinations of them.

**Methods**

The methods are divided into 2 parts: A) description of the system that generates interactive display of the treatments for a user-chosen treatment; and B) system evaluation.

## A. System Architecture

The system to automatically generate treatment displays is built as a pipeline that combines the following informatics tools and methods: 1) a query processing tool using UMLS Metathesaurus[7] for extracting concepts, 2) Semantic Medline Database[8] for extracting treatment relationships, 3) a term-frequency based algorithm to eliminate non-specific treatment names, 4) a normalized-frequency based ranking of the remaining treatments, and 5) a graphical interface to interactively learn about the treatments. Figure 1 depicts the system architecture and flow.

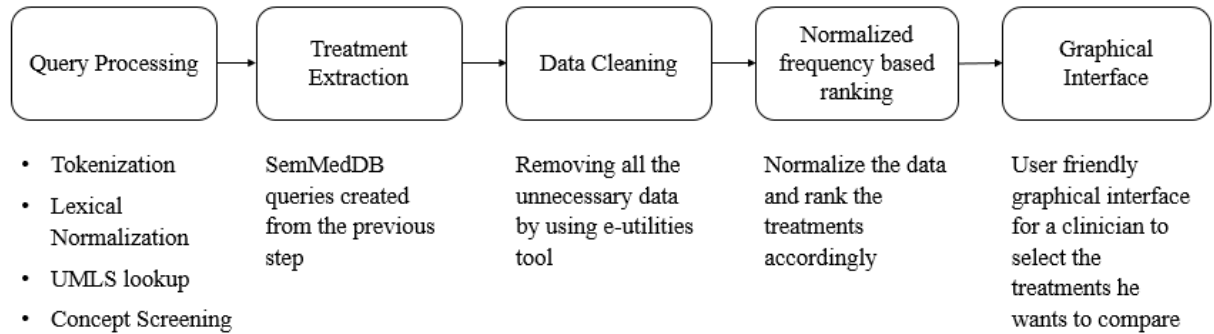

**Figure 1.** System Architecture

### 1. Query Processing

As shown in Figure 1, the query for a clinical topic (e.g., congestive heart failure) is initially processed with a UIMA-based (Unstructured Information Management Architecture[9]) concept extraction pipeline that maps concepts in narrative text to the UMLS Metathesaurus. The components in the pipeline are:

- Tokenization or splitting the query into individual token – adapted from the OpenNLP suite[10];
- Lexical normalization or converting words to a canonical form – the Lexical Variant Generation (LVG) terminology from the UMLS metathesaurus is compressed by i) converting the terms to lowercase; ii) removing the terms where the normalized word has more than one token; and iii) removing the terms that have the same base form.
- UMLS Metathesaurus lookup – performed using a well-known efficient algorithm called Aho-Corasick string matching algorithm[11] that loads the normalized tokens and their substrings as the individual states of the corresponding finite state machine, where the transitions between the different states represent the different terms formed by the original tokens.
- Concept screening – select only the UMLS concepts in the query that are members of the semantic groups disorder (Abnormality, Dysfunction, Disease or Syndrome, Finding, Injury or Poisoning, Pathologic Function, and Sign or Symptom[12]).

### 2. Treatment extraction from Medline abstracts

The treatments for the disorder extracted in the previous step is obtained by querying the Semantic Medline Database. Semantic Medline Database (SemMedDB)[8] is a repository of semantic predications (subject-predicate-object triples) extracted by SemRep[13], a semantic interpreter of biomedical text. The query we use searches SemMedDB for sentences where the object of the sentence is related to disorder and the predicate of the sentence is 'TREATS' (e.g. radiofrequency ablation treats atrial fibrillation). Algorithm 2 of our previous work[14] details how the queries are created to retrieve the relevant semantic predications. Additionally the query is constrained to be novel which eliminates predications that have a generic argument as determined by hierarchical depth in the Metathesaurus[2]. The query then yields a set of treatments for the disorder (example: congestive heart failure), mentioned in Medline abstracts.

### 3. Eliminating non-specific treatment names

When treatments were extracted from the Semantic Medline Database, many broadly named procedures like 'Therapeutic Procedure' and 'Adjustment Action' were also extracted. These procedures, being unimportant to a clinician, had to be removed while displaying keeping the relevant treatments. Some of them were removed using a feature present in SemMedDB that prunes concepts according to their relative granularity in the concept hierarchy. In addition, we also calculated the number of abstracts in which the combination of procedure (e.g. 'Adjustment Action') and disorder (e.g. 'Congestive Heart Failure') appears in Medline compared with the number of abstracts in which the treatment itself appears in Medline. The treatments, for which the former number was relatively low compared to the later number (< 1%) was categorized as a broadly named procedures and hence these procedures were removed from the data set.

### 4. Normalized-frequency based ranking

There are multiple ways in which the treatments might be ordered. We chose to first normalize the treatments by their presence in Medline abstracts across individual year ranges (1980-1985, 1986-1990, 1991-1995, 1996-2000, 2001-2005, 2006-2010, and 2011-2013) and construct a linear combination of the these normalized frequencies with one combination giving more weight to the recent years, the other giving equal weights and also providing the option in the interface (discussed in next section) for the clinician user to choose their own weights.

Normalization across individual year ranges was done by using the formula given below.

$$Normalized\ (e_i) = \frac{e_i - E_{min}}{E_{max} - E_{min}}$$

where, $E_{min}$ represents the minimum value across the individual year range; $E_{max}$ represents the maximum value across the individual year range.

If a clinician is interested in treatments that are often used in recent years, we rank the treatments by giving weights of 1, 2, 3, 4, 5, 6, and 7 for the respective year-ranges. This would result in ranking of the treatments in which more importance is given to the treatments that appeared in the recent years. The first of our weighing method was called 'New Treatments Weighing Method'. In the 'New Treatments Weighing Method', we gave weights of 1, 2, 3, 4, 5, 6, and 7 to the year ranges. This will result in ranking of the treatments in which the treatments that appeared recently are given more importance. The second weighing method was called 'Established Treatments Weighing Method'. In the 'Established Treatments Weighing Method', we gave equal weights to all the year ranges, i.e. weights of 1 are given to all the year ranges. This weighing method results in ranking of the treatments in which all the treatments that appeared are given equal importance.

### 5. Graphical interface

While building the interface, we tried to convey simplicity and provide a reasonable starting point for clinicians to begin their information gathering for a given disease treatment.

Building the first iteration of the tool, we started by identifying use cases a clinician would have when starting to gather treatment information for a particular disease. In the first use case we identified, clinicians search for novel disease treatments in scientific journals. In the second use case, clinicians search for established disease treatments in scientific journals. In the third use case, clinicians search are more interested in treatments studied during a particular timespan more than the others. Finally, in the fourth use case, the clinician would compare the number of individual disease treatments mentions with their intersections to find popular combinations of treatments.

**B. System Evaluation**

For the evaluation study, we used a case study approach by assessing the output of the knowledge summary system on the treatment of two conditions: *atrial fibrillation* (a topic relatively well understood) and *congestive heart failure* (a topic less covered, but is an increasingly complex vast field with knowledge from huge literature). For atrial fibrillation, we compared the rankings of the system with the practice guideline – Management of Patients With Atrial Fibrillation (Compilation of 2006 ACCF/AHA/ESC and 2011 ACCF/AHA/HRS Recommendations).[15] For atrial fibrillation, we compared the rankings of the system with the practice guideline – 2013 ACCF/AHA Guideline for the Management of Heart Failure: Executive Summary: A Report of the American College of Cardiology Foundation/American Heart Association Task Force on Practice Guidelines.[16]

For both these disorders, we report the precision, recall and f-score values for top-10 treatments, top-20 treatments, and so on until top-100 treatments. Precision is the ratio of the number of treatments in the guideline and extracted by the system and the number of all the treatment extracted by the system. Recall is the ratio of the number of treatments in the guideline and extracted by the system and the number of all the treatments in the gold standard. F-score is the harmonic mean of precision and recall.

**Results**

**Atrial Fibrillation**

After we extracted the treatments for 'Atrial Fibrillation' from SemMedDB, 623 treatments were mined. Out of these, several treatments were generalized procedures which would not be of any use to a clinician. By using e-utilities tool and a threshold frequency of 1% we cleaned the data of these non-specific treatments. On performing the cleaning, 360 of these were removed and 263 treatments remained. On these 263 treatments, we used our two ranking methods explained in the methods section.

New Treatments Weighing method: The top 5 treatments that were generated using this method along with their UMLS Concept Unique Identifiers are Ablation (C0547070), Electric Counter shock (C0013778), anticoagulation (C0003281), Stroke prevention (C1277289), and Cardiac ablation (C0162563). The precision-recall curve for this weighing method for the top 100 treatments and the F-score values are displayed in Figure 2 and Table 1.

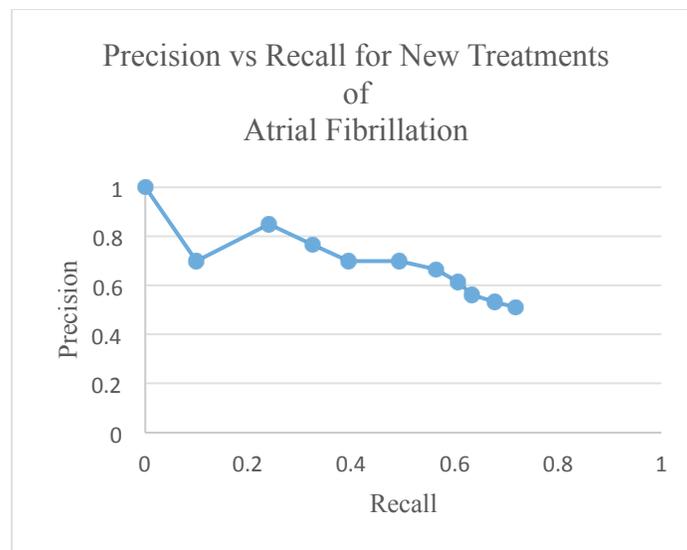

**Figure 2.** Precision vs Recall for New Treatments of Atrial Fibrillation

**Table 1.** Recall, Precision and F-scores for New Treatments of Atrial Fibrillation

| Number of treatments | Recall | Precision | F-score |
|---|---|---|---|
| 1 | 0.000 | 1.000 | 0.000 |
| 10 | 0.099 | 0.700 | 0.173 |
| 20 | 0.239 | 0.850 | 0.374 |
| 30 | 0.324 | 0.767 | 0.455 |
| 40 | 0.394 | 0.700 | 0.505 |
| 50 | 0.493 | 0.700 | 0.579 |
| 60 | 0.563 | 0.667 | 0.611 |
| 70 | 0.606 | 0.614 | 0.610 |
| 80 | 0.634 | 0.563 | 0.596 |
| 90 | 0.676 | 0.533 | 0.596 |
| 100 | 0.718 | 0.510 | 0.596 |

Established Treatments Weighing method: The top 5 treatments that were generated using this method are Electric Counter shock (C0013778), Ablation (C0547070), anticoagulation (C0003281), Amiodarone (C0002598), and Stroke prevention (C1277289). The precision-recall curve for this weighing method for the top 100 treatments and the F-score values are displayed in Figure 3 and Table 2.

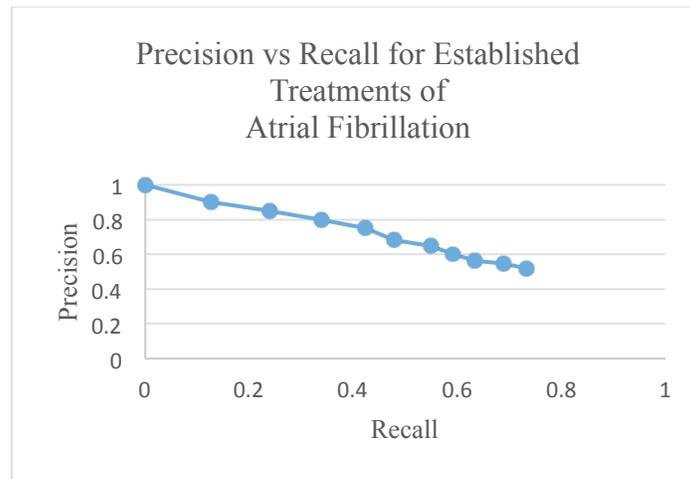

**Figure 3.** Precision vs Recall for Established Treatments of Atrial Fibrillation

**Table 2.** Recall, Precision and F-scores for Established Treatments of Atrial Fibrillation

| Number of Treatments | Recall | Precision | F-Square |
|---|---|---|---|
| 1 | 0.000 | 1.000 | 0.000 |
| 10 | 0.127 | 0.900 | 0.222 |
| 20 | 0.239 | 0.850 | 0.374 |
| 30 | 0.338 | 0.800 | 0.475 |
| 40 | 0.423 | 0.750 | 0.541 |
| 50 | 0.479 | 0.680 | 0.562 |
| 60 | 0.549 | 0.650 | 0.595 |
| 70 | 0.592 | 0.600 | 0.596 |
| 80 | 0.634 | 0.563 | 0.596 |
| 90 | 0.690 | 0.544 | 0.609 |
| 100 | 0.732 | 0.520 | 0.608 |

**Congestive Heart Failure**

After we extracted the treatments for 'Congestive Heart Failure' from SemMedDB, 1,608 treatments were mined. Out of these, several treatments were generalized procedures which would not be of any use to a clinician. On performing the cleaning, 1,355 of these were removed and 253 treatments remained. On these 253 treatments, we used our two ranking methods explained in the methods section.

New Treatments Weighing method: The top 5 treatments that were generated using this method are Angiotensin-Converting Enzyme Inhibitors (C0003015), Adrenergic beta-Antagonists (C0001645), Cardiac resynchronization therapy (C1167956), Brain natriuretic peptide (C0054015), and Digoxin (C0012265). The precision-recall curve for this weighing method for the top 100 treatments along with the F-score values are displayed below in Figure 4 and Table 3.

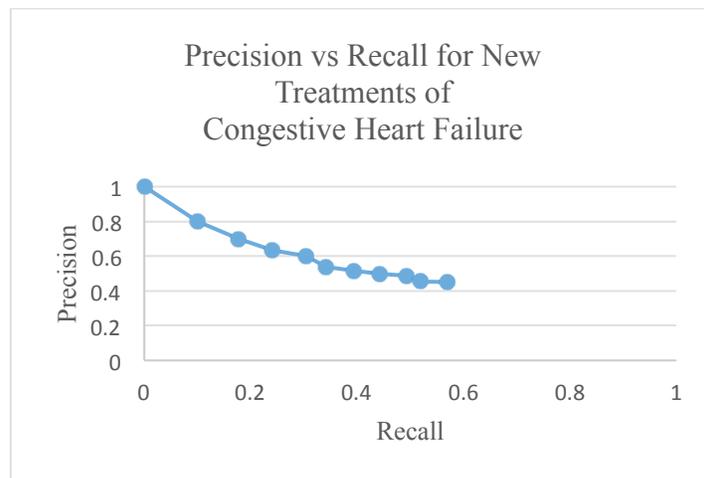

**Figure 4.** Precision vs Recall for New Treatments of Congestive Heart Failure

**Table 3.** Recall, Precision and F-scores for New Treatments of Congestive Heart Failure

| Number of Treatments | Recall | Precision | F-Square |
|---|---|---|---|
| 1 | 0.000 | 1.000 | 0.000 |
| 10 | 0.101 | 0.800 | 0.180 |
| 20 | 0.177 | 0.700 | 0.283 |
| 30 | 0.241 | 0.633 | 0.349 |
| 40 | 0.304 | 0.600 | 0.403 |
| 50 | 0.342 | 0.540 | 0.419 |
| 60 | 0.392 | 0.517 | 0.446 |
| 70 | 0.443 | 0.500 | 0.470 |
| 80 | 0.494 | 0.488 | 0.491 |
| 90 | 0.519 | 0.456 | 0.485 |
| 100 | 0.570 | 0.450 | 0.503 |

Established Treatments Weighing method: The top 5 treatments that were generated using this method are Angiotensin-Converting Enzyme Inhibitors (C0003015), Adrenergic beta-Antagonists (C0001645), Digoxin (C0012265), Cardiac resynchronisation therapy (C1167956), and Milrinone (C0128513). The precision-recall curve for this weighing method for the top 100 treatments along with the F-score values are displayed below in Figure 5 and Table 4.

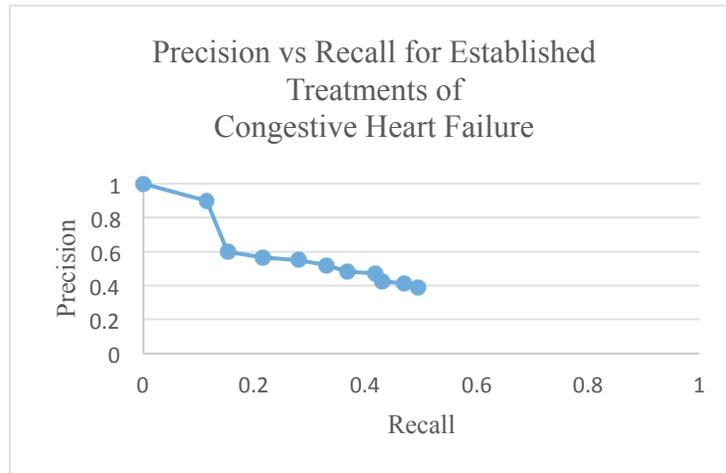

**Figure 5.** Precision vs Recall for Established Treatments of Congestive Heart Failure

**Table 4.** Recall, Precision and F-scores for Established Treatments of Congestive Heart Failure

| Number of Treatments | Recall | Precision | F-Square |
|---|---|---|---|
| 1 | 0.000 | 1.000 | 0.000 |
| 10 | 0.114 | 0.900 | 0.202 |
| 20 | 0.152 | 0.600 | 0.242 |
| 30 | 0.215 | 0.567 | 0.312 |
| 40 | 0.278 | 0.550 | 0.370 |
| 50 | 0.329 | 0.520 | 0.403 |
| 60 | 0.367 | 0.483 | 0.417 |
| 70 | 0.418 | 0.471 | 0.443 |
| 80 | 0.430 | 0.425 | 0.428 |
| 90 | 0.468 | 0.411 | 0.438 |
| 100 | 0.494 | 0.390 | 0.436 |

**Interface**

For using the tool, a clinician would follow the steps outlined below (See Figure 6 for a screenshot):

1. Clinician searches for a disease they want to gather more information about either by entering the disease name or the UMLS Concept Unique Identifier.
2. Clinician selects a treatment trend their interested in (Novel, Consistent, Custom)
3. Clinician sorts treatments (Trend matches: most to least, Treatment matches: least to most)

4. Clinician selects treatments they would like to see graphed by the number of mentions over time.

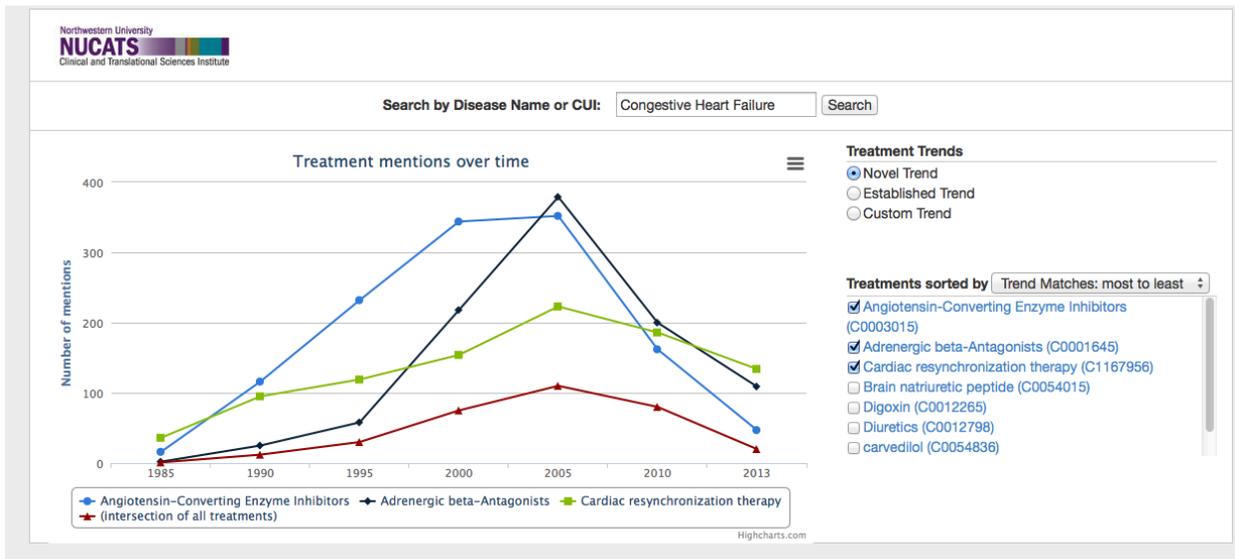

**Figure 6.** Web interface for showing treatment trends

**Discussion**

Atrial Fibrillation: The precision and recall curve shows that as the recall increases, precision decreases. In a web-interface, one needs to display enough number of treatments such that both the precision and recall are moderately high. For the case of New Established Treatments method, the f-score of 0.611 is highest at precision of 0.667. The recall at this precision is 0.611. This cutoff point is reached at 60 treatments. This cutoff point is also applicable to the Established Treatments method. At 60 treatments, the f-score is 0.595, which is approximately as high as the largest f-score value, 0.609 for this method.

Congestive Heart Failure: For the case of New Treatments method, the best cutoff point for f-score is 0.4915, which occurs at 80 treatments. Even though the highest f-score is 0.503 (which occurs at 100 treatments), 80 treatments would be a better cutoff since there is not much difference in the F-scores between these two and selecting the 100 treatments would mean just going through more number of treatments. The recall and precision at this cutoff are 0.494 and 0.488 respectively, which are moderately high. For the case of Established Treatments method, we can observe the maximum f-score (0.443) is reached at 70 treatments. The recall and precision values at this cutoff are 0.418 and 0.471 respectively.

For the two disorders we have discussed, we can observe that a cutoff of 70 treatments would result in a moderately high precision and recall. In our interface, we are able to display as many treatments as there are for the disorder; however, text mining based interfaces that display fewer treatments might be missing few important treatments.

**Limitations**

The analysis was performed on only two diseases 'Atrial Fibrillation' and 'Congestive Heart Failure'. Although this might be sufficient to generalize the system for cardiology, more experiments are needed to generalize it for all the disorders.

The threshold value of 1% that was used as a part of data cleaning is arbitrary decided by observing various data sets. Even though this value was helpful in removing several generalized procedures, few generalized procedures such as 'Hospitalization' and 'Secondary Prevention' are still not filtered. In the future, we will incorporate this ratio threshold values and will perform analysis on treatment procedures of various diseases and come up with an optimized value for this.

Another limitation is that SemRep like any text mining system misses few relationships (resulting in lower recall for our system) and falsely assigns relationships (resulting in lower precision). In a previous study, the average precision of SemRep for predications on the treatment of four diseases was 73% (recall not calculated as it was difficult to determine the false negatives).[17] Moreover, SemRep operates at a sentence-level, which means that when there is an implicit relationship between treatments and a disease (such as the disease mentioned in one sentence and the treatment mentioned using an anaphora to the disease in another), the system would not be able to extract it.

When extracting the treatments, the same treatments are sometimes repeated with different names. For example, 'anti-Coagulation' and 'Anti-Coagulation therapy' refers to the same treatment, but they are being listed separately because of the challenges with normalization of these named-entities. Another example is that of 'Warfarin' and 'Warfarin Therapy'. A related example is 'Defibrillator' and 'Cardioversion' which are synonyms according to UMLS Metathesaurus. The practice guidelines only used the term 'Cardioversion' and in our extracted treatment names 'Defibrillator' and 'Cardioversion' are listed separately. Because the same treatment is being displayed under different names, the data for the number of times a treatment has appeared will not be accurate and the clinician user will also have a suboptimal experience.

**Conclusion and Future work**

For Atrial Fibrillation disorder, maximum f-score for the New Treatments weighing method is 0.611, which occurs at 60 treatments. For Congestive Heart Failure disorder, maximum f-score for the New Treatments weighing method is 0.503, which occurs at 100 treatments. By observing the precision, recall and f-scores of the ranking methods we proposed, we can infer that the extraction and ranking schema we proposed might be a good solution for displaying the treatment procedures. The analysis we have done was only on two diseases, 'Atrial Fibrillation' and 'Congestive Heart Failure'. In future, we will be performing similar kind of analysis on more number of diseases. By analyzing these results, we target to get a generalized optimal solution for other disorders. We discussed various limitations that reduced the accuracy of the results generated. In future studies, we aim to come up with solutions to these limitations.

**Acknowledgements**

This work was made possible by funding from the National Library of Medicine K99/R00 LM011389.